\documentclass[acmtog, authorversion,screen]{acmart}

\AtBeginDocument{
  \providecommand\BibTeX{{
    \normalfont B\kern-0.5em{\scshape i\kern-0.25em b}\kern-0.8em\TeX}}}

\setcopyright{acmcopyright}
\copyrightyear{2023}
\acmYear{2023}
\acmDOI{}

\acmConference[CausalStats'23]{When Causal Inference meets Statistical Analysis}{April 17-21, 2023}{Paris, France}

\acmBooktitle{CausalStats'23: When Causal Inference meets Statistical Analysis, April 17-21, 2023, Paris, France} 

\usepackage{amsmath,bm}
\usepackage{multirow}
\usepackage{algorithm}

\def\x{{\mathbf x}}
\def\y{{\mathbf y}}
\def\q{{\mathbf q}}
\def\r{{\mathbf r}}
\def\A{{\mathbf A}}

\def\C{{\mathbf C}}
\def\H{{\mathbf H}}
\def\R{{\mathbf R}}
\def\Q{{\mathbf Q}}

\def\V{{\mathbf V}}

\def\Z{{\mathbf Z}}
\def\m{{\mathbf m}}
\def\z{{\mathbf z}}
\def\P{{\mathbf P}}
\def\S{{\mathbf S}}
\def\Id{{\textbf{Id}}}
 
\newcommand{\Sigmab}{{\bm \Sigma}}
\newcommand{\Phib}{{\bm \Phi}}

\usepackage{color}

\usepackage{hyperref}
\hypersetup{
    unicode=false,          
    pdftoolbar=true,        
    pdfmenubar=true,        
    pdffitwindow=false,     
    pdfstartview={FitH},    
    pdftitle={},         
    pdfauthor={}, 
    pdfsubject={} 
    pdfcreator={}           
    pdfproducer={},         
    pdfkeywords={},         
    pdfnewwindow=true,      
    colorlinks=true,        
    linkcolor={blue},       
    citecolor=blue,         
    filecolor=blue,         
    urlcolor=blue           
}

\usepackage{algpseudocode}

\begin{document}

\title{Graphs in State-Space Models for Granger Causality in Climate Science}

\author{V\'ictor Elvira}
\email{victor.elvira@ed.ac.uk} \orcid{0000-0002-8967-4866}
\affiliation{%
\institution{University of Edinburgh}
  \city{Edinburgh}
  \country{UK}
  \postcode{}
}
\author{\'Emilie Chouzenoux}
\email{emilie.chouzenoux@centralesupelec.fr} \orcid{0000-0003-3631-6093}
\affiliation{%
\institution{Inria Saclay}
  \city{Paris}
  \country{France}
  \postcode{}
}
\author{Jordi Cerd\`a}
\email{jordi.cerda@uv.es} \orcid{0000-0003-4512-6005} \affiliation{%
\institution{Universitat de Val\`encia}
  \city{Paterna}
  \country{Spain}
  \postcode{46980}
}
\author{Gustau Camps-Valls}
\email{gustau.camps@uv.es} \orcid{0000-0003-1683-2138}
\affiliation{%
\institution{Universitat de Val\`encia}
  \city{Paterna}
  \country{Spain}
  \postcode{46980}
}
  
\renewcommand{\shortauthors}{Elvira, Chouzenoux, Cerd\`a, and Camps-Valls}

\begin{abstract}
  Granger causality (GC) is often considered not an actual form of causality. Still, it is arguably the most widely used method to assess the predictability of a time series from another one. Granger causality has been widely used in many applied disciplines, from neuroscience and econometrics to Earth sciences. 
  We revisit GC under a graphical perspective of state-space models. For that, we use GraphEM, a recently presented expectation-maximisation algorithm for estimating the linear matrix operator in the state equation of a linear-Gaussian state-space model. Lasso regularisation is included in the M-step, which is solved using a proximal splitting Douglas-Rachford algorithm. Experiments in toy examples and challenging climate problems illustrate the benefits of the proposed model and inference technique over standard Granger causality methods.
\end{abstract}



\maketitle

\section{Introduction}
\label{sec:intro} 

Granger causality (GC)~\cite{granger_investigating_1969} is undoubtedly the most widely used method to infer not causal but {\em predictability} relations from observational time series, and is the common tool in Earth sciences~\cite{Runge19natcom}, neurosciences~\cite{reid2019advancing}, social sciences~\cite{marini1988causality} and finance~\cite{hicks1980causality}. Granger causality tests whether one time series is predictive of another beyond what can be explained by the other series alone. GC implicitly tells us about the amount of {\em information} present in the first time series that is not in the second time series and is helpful for its {\em forecast}. Other methods rely on similar concepts of {\em information flow} and {\em predictability}: connections can be established between GC and transfer entropy \cite{schreiber_measuring_2000}, directed information \cite{massey1990causality},
convergent cross-mapping \cite{sugihara_detecting_2012}, Liang's measure of information flow \cite{liang_information_2015}, and with the 
graphical causal model perspective \cite{white_linking_2011}. Recent structural equation models (SEM) can also be seen as particular instantiations of GC: the time series Linear non-Gaussian acyclic model (TS-LiNGAM) \cite{hyvarinen_estimation_2010} can be viewed as a vector autoregressive (VAR) model with the non-Gaussian variable, and the time series models with independent noise (TiMINo) \cite{peters_causal_2013} can be seen as a non-linear-extension of a VAR model. These models require autoregressive modelling to estimate the variance of residuals, which makes GC struggle when the process has been filtered, down-sampled or observed with additive noise \cite{barnett2015granger}, which are common situations in many fields of science and engineering. Also, finite-order VAR models require choosing the order and do not allow incorporating latent states that may not be observed.

This work focuses on state-space models (SSMs) to identify the causal structure. SSMs model a hidden state evolving in a Markovian manner over discrete time. A sequence of observations is acquired, each being a transformed, partial, and noisy version of the hidden state.  
A common task is estimating the hidden state conditioned on available observations (Bayesian filtering). The task of estimating each state using the whole sequence of observations is called Bayesian smoothing. The linear-Gaussian SSM is widely used, and exact computation of filtering and smoothing distributions is possible via the Kalman filter and the Rauch-Tung-Striebel (RTS) smoother \cite[Chapter 8]{Sarkka}. Static parameter estimation is possible using expectation-minimisation (EM) or optimisation-based methods \cite[Chapter 12]{Sarkka}. 

We here review the GraphEM method \cite{elvira2022graphem} for estimating the linear matrix of a linear-Gaussian state-space model's state equation and evaluating its performance for GC in synthetic and real-life experiments in the Earth and climate sciences. 
Estimating such a matrix allows us to obtain valuable information about the hidden process, not only for inference purposes but also for understanding the uncovered relations among the state dimensions, in the line of graphical modelling methods for time series~\cite{Eichler2012,Bach04,Barber10}. 

GraphEM estimates directed graphs representing causal dependencies between states of a multi-dimensional state containing uni-dimensional time series, with a sparsity constraint in the linear matrix. It applies the EM framework for MAP estimation, with the E-step using Kalman filtering and RTS smoothing and the M-step solving a Lasso-like problem through Douglas-Rachford proximal splitting \cite{CombettesDR}. Experiments on toy and real climate problem simulations show fast, practical convergence, and better performance than state-of-the-art methods.

\section{BACKGROUND}\label{sec:format}

Let us consider the following Markovian SSM,
 \begin{equation}
   \begin{array}{ll}
       \x_{k} &= \A \x_{k-1} + \q_{k},   \\
     \y_k &= \H \x_k + \r_{k},
    \end{array}
    \label{eq:model}
    \end{equation}
where, for $k=1,\ldots,K$, $\x_{k}\in \mathbb{R}^{N_x}$ is the hidden state at time $k$,  $\y_{k}\in \mathbb{R}^{N_y}$ is the associated observation, $\A \in \mathbb{R}^{N_x \times N_x}$, $\H \in \mathbb{R}^{N_y \times N_x}$, $\{ \q_k \}_{k=1}^K \sim \mathcal{N}(0,\Q)$ is the i.i.d. state noise process and $\{ \r_k \}_{k=1}^K \sim \mathcal{N}(0,\R)$ is the i.i.d. observation noise process. The state process is initialized as $\x_0 \sim \mathcal{N}(\x_0 ; \bar \x_0, \P_0)$ with known $\bar \x_0$ and $\P_0$. 

The interest in the linear-Gaussian model of  \eqref{eq:model} resides on computing the filtering distributions $p(\x_k|\y_{1:k})$ and the smoothing distributions $p(\x_k|\y_{1:K})$, with $\y_{1:k} = \{ \y_j \}_{j=1}^k$. The Kalman filter provides exact forms of these as $p(\x_k|\y_{1:k}) = \mathcal{N}(\x_k|\m_k,\P_k)$~\cite{Kalman60}. The RTS smoother also yields exact smoothing distributions, $p(\x_k|\y_{1:K}) = \mathcal{N}(\x_k|\m_k^s,\P_k^s)$. To use either algorithm, the exact model parameters $\A$, $\H$, $\Q$, and $\R$ must be known~\cite{Sarkka}. In the next section, we introduce GraphEM \cite{elvira2022graphem} for the joint estimation of the unknown $\A$ and the hidden states.
 
\section{The GraphEM APPROACH}
\label{sec:proposed}

\subsection{Role of sparsity and connection to GC}

GraphEM approach, introduced in \cite{elvira2022graphem,chouzenoux2020} provides a MAP estimation of matrix $\A$ 
 using a sparsity prior knowledge. Matrix $\A$ describes the relations among the dimensions of the hidden state, that is, the interactions of the state dimensions between consecutive time steps. A sparse transition matrix $\A$ allows us to interpret the multi-variate process under Granger causality (also called predictive causality) on the process $\x_k$ (i.e., not in the observed time series, unlike in standard Granger causality). The $(n,m)$ entry in the $\A$ encodes the weight
in which the $m$th time series in the hidden state affects the $n$th time series in the next time step, being zero if it does not have
any effect. A null entry can be understood as if the $m$th time series does not bring any information to predict the $n$th time series (given the others). 

\subsection{EM methodology}

GraphEM seeks the maximum of the MAP function $p(\A|\y_{1:K})$ $\propto$ $p(\A)p(\y_{1:K}|\A)$, i.e., at solving
\begin{equation}
    \text{minimise}_{\A \in \mathbb{R}^{N_x \times N_x}} (\varphi_K(\A) = - \log p(\A) 
    - \log p(\y_{1:K}| \A)).
\end{equation}
In order to promote sparsity in the estimation of the $\A$ matrix, the regularisation function $\varphi_0(\A) = - \log p(\A)$ is set as
\begin{equation}
    (\forall \A \in \mathbb{R}^{N_x \times N_x}) \quad \varphi_0(\A) = \gamma \| \A \|_1,
\end{equation}
with $\gamma > 0$ a penalisation weight. The $\ell_1$ penalty, also known as Lasso, is convex and, as such, can be efficiently handled through the proximal optimisation method.

    The direct minimisation of $\varphi_K$ is difficult due to the intricate form of the likelihood term \cite{Sarkka}:
    \begin{equation}
        \log p(\y_{1:K}| \A) = \sum_{k=1}^K \frac{1}{2} \log | 2 \pi \S_k(\A)| + \frac{1}{2} \z_k(\A)^\top \S_k(\A)^{-1} \z_k(\A),
    \end{equation}
with $\z_k(\A) = \y_k - \H\A\m_{k-1}(\A)$  and $\S_k(\A)$ the covariance matrix of the predictive distribution $p(\y_{k}|\y_{1:k-1})= \mathcal{N}\left(\y_{k};\H\A\m_{k-1}(\A),\S_{k}(\A)\right)$, both being sided products of the Kalman filter ran for a given $\A$ (see \cite[Section 4.3]{Sarkka}). 

The EM approach adopted in GraphEM allows deriving an upper bound that is more tractable and, thus, easier to minimise. As shown in \cite{elvira2022graphem}, the following function, parameterised by $\A'$, majorises the MAP objective function $\varphi_K$ for every $\A \in \mathbb{R}^{N_x \times N_x}$:
\begin{multline}
   \mathcal{Q} (\A;\A')  =  \frac{K}{2} \text{tr} \left(\Q^{-1} (\Sigmab - \C \A^\top - \A \C^\top + \A \Phib \A^\top) \right)  + \varphi_0(\A) + \mathcal{C}
   \label{eq:majQ}
\end{multline}
where $\rm{tr}$ is the trace operator, and $\mathcal{C}$ is a constant term independent from~$\A$. The GraphEM algorithm is then deduced, following a majorisation-minimisation principle \cite{jacobson} as described in Alg.~\ref{algo:EM_MAP}. The inner problem involved in the M-step is a convex Lasso-like problem that is solved through Douglas-Rachford iterations \cite{CombettesDR} given in the next Section. The convergence of Alg.~\ref{algo:EM_MAP} is established in \cite[Th. 1]{elvira2022graphem}.

 \begin{algorithm}[h!]
\begin{flushleft}
 \noindent Initialisation of $\A^{(0)}$\\
 \noindent For $i = 1,2,\ldots$\\
     $\phantom{aaa}$ \textbf{(E-step)} Run the Kalman filter and RTS smoother, $\A':=\A^{(i-1)}$\\
     \hspace{1.7cm}Construct $\mathcal{Q}(\A;\A^{(i-1)})$ with \eqref{eq:majQ}.\\    $\phantom{aaa}$ \textbf{(M-step)} Update $\A^{(i)} = \text{argmin}_{\A} \left( \mathcal{Q}(\A;\A^{(i-1)})\right)$ (see Sec.~\ref{sec_m_step})
        \caption{GraphEM algorithm.}
     \label{algo:EM_MAP}
\end{flushleft}
\end{algorithm}

\subsection{Computation in the M-step}
\label{sec_m_step}
At an iteration $i \in \mathbb{N}$, the M-step minimises the function $\mathcal{Q}(\A ;\A^{(i-1)})$, obtained by plugging $\A':=\A^{(i-1)}$ in Eq.\eqref{eq:majQ}. This minimisation problem, sometimes called Lasso regression~\cite{Tibshirani}, has been much studied in the literature of optimisation \cite{Schmidt,Bach}, and most of the methods proposed to solve it rely on the proximity operator \cite{Combettes2010,Bauschke}.\footnote{See also \url{http://proximity-operator.net/}} 

Let us decompose, for every $\A \in \mathbb{R}^{N_x \times N_x}$, $\mathcal{Q}(\A ;\A^{(i-1)}) = f_1(\A) + f_2(\A)$, where $f_1(\A) = \frac{K}{2}  \text{tr} \left(\Q^{-1} (\Sigmab - \C \A^\top - \A \C^\top + \A \Phib \A^\top) \right)$, and $f_2 = \varphi_0$. GraphEM relies on the Douglas-Rachford (DR) algorithm to minimise $f_1 + f_2$; DR is a fixed-point proximal strategy for convex optimisation that benefits from sound convergence guarantees \cite{CombettesDR} and has demonstrated its great practical performance in matrix optimisation problems related to graphical inference applications~\cite{Benfenati18}. 

DR iterations are summarised in Alg.~\ref{algo:DR}, in the simplified case when $\Q = \sigma_{\Q}^2 \bf{Id}$ (see general case in \cite{chouzenoux2020}). DR generates a sequence $\{\A_j\}_{j \in \mathbb{N}}$ guaranteed to converge to a minimiser of $f_1 + f_2$. In practice, for the iteration $i$ in GraphEM, we run the DR method with $\theta = 1$ and initialisation $\Z_0 = \A^{(i-1)}$, i.e. the majorant function tangency point. Moreover, we stop the DR loop as soon as $|(f_1 + f_2)(\A_{j+1}) - (f_1 + f_2)(\A_{j})| \leq \varepsilon$ (typically, $\varepsilon = 10^{-3}$), the DR output $\A_j$ being used for defining the GraphEM iterate $\A^{(i)}$.

 \begin{algorithm}
\begin{flushleft}
 Set $\Z_0 \in \mathbb{R}^{N_x \times N_x}$ and $\theta \in (0,2)$\\
 For $j =1,2,\ldots$\\
 $\phantom{aaa}$ $\A_j = \left(\text{sign}((\Z_j)_{nm}) \times\max(0,|(\Z_j)_{nm}| - \theta)
     \right)_{1 \leq n,m \leq N_x}$\\
 $\phantom{aaa}$ $\V_j = 
  \left(\frac{\theta K}{\sigma_\Q^2} \C+2 \A_j - \Z_j\right) \left(\frac{\theta K}{\sigma_\Q^2} \Phib + \Id\right)^{-1}$\\
 $\phantom{aaa}$ $\Z_{j+1} = \Z_j + \theta (\V_j - \A_j)$
\end{flushleft}
 \caption{Douglas-Rachford algorithm for M-step.}
\label{algo:DR}
\end{algorithm} 

\section{EXPERIMENTAL EVALUATION}
\label{sec:experiment}

\subsection{Synthetic examples}

We generate synthetic data using \eqref{eq:model} with $N_x = N_y$. $\A$ is composed of $b$ blocks of dimensions $(b_j)_{1 \leq j \leq b}$ such that $N_x = \sum_{j=1}^b b_j$. Each diagonal block is randomly generated, and a projection with a maximum singular value of less than one is applied for stability. $\H, \Q, \R, \P_0$ are set to $\rm{\Id}$, $K = 10^3$, $\sigma_\Q^2 \rm{\Id}$, $\sigma_\R^2 \rm{\Id}$, $\sigma_\P^2 \rm{\Id}$ respectively. Four datasets were created (Table~\ref{tab:data}), and performance was assessed using RMSE on $\A$ and precision, recall, specificity, accuracy, and F1 score for graph edge detection (the elements of the estimated matrix are clipped if smaller than $10^{-10}$ in absolute value).

\begin{table}[h]
\centering
{\small
\begin{tabular}{|c||c|c|}
\hline
Dataset & $(b_j)_{1 \leq j \leq b}$ & $(\sigma_\Q,\sigma_\R,\sigma_\P)$\\
\hline\hline
A & $(3,3,3)$ & $(10^{-1},10^{-1},10^{-4})$\\
\hline
B & $(3,3,3)$ & $(1,1,10^{-4})$\\
\hline
C & $(3,5,5,3)$ & $(10^{-1},10^{-1},10^{-4})$\\
\hline
D & $(3,5,5,3)$ & $(1,1,10^{-4})$\\
\hline
\end{tabular}
}
\caption{Description of datasets}
\label{tab:data}
\end{table}

For each dataset, we ran the GraphEM algorithm with the stopping rule $|\varphi_K(\A^{(i)}) - \varphi_K(\A^{(i-1)})| \leq 10^{-3}$. A maximum number of $50$ iterations was set, and it was always sufficient to reach this criterion. The algorithm is initialized by setting a random and dense matrix $\A^{(0)}$, ensuring the stability of the process. Parameter $\gamma$, balancing the weight of the sparse prior, is optimized on a single realisation thanks to a manual grid search to maximize the accuracy score. We also provide the ML estimator's results, computed using the EM algorithm (MLEM). In this case, the M-step has a closed-form solution~\cite[Th.12.5]{Sarkka}. In addition, we compare two Granger-causality approaches \cite{Bressler2011} for graphical modelling. The first algorithm, pairwise Granger Causality (PGC) explores the $N_x(N_x$-$1)$ possible dependencies among two nodes, at each time independently from the rest. The second approach, the conditional Granger Causality (CGC), operates similarly. Still, for each pair of nodes, it also considers the information of the other $N_x-2$ signals to evaluate whether one node provides information to the other when the rest of the signals are observed. We also do a manual grid search for fine-tuning the parameters of both PGC and CGC (more information can be found in \cite{Luengo19}). PGC and CGC do not estimate a weighted graph but a binary one, so RMSE is not calculated in those cases. 

\begin{figure}
\begin{tabular}{cc}
\includegraphics[width = 0.2\textwidth]{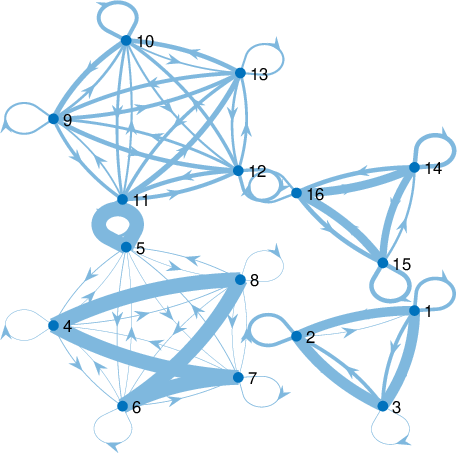} & \includegraphics[width = 0.2\textwidth]{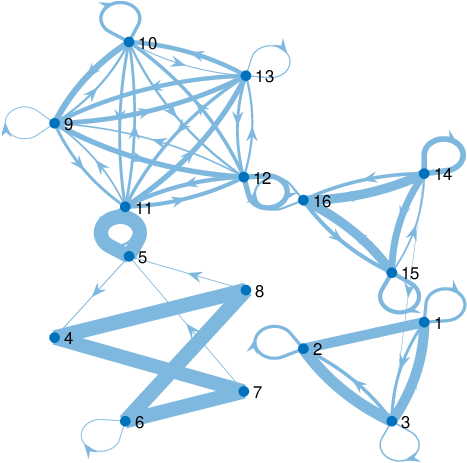} 
\end{tabular}
\vspace*{-0.2cm}
\caption{True graph (left) and GraphEM estimate (right) for dataset C.}
\label{fig:datasetC}
\end{figure}

The results averaged on $50$ realisations are presented in Table~\ref{tab:results}. MLEM does not promote sparsity in the graph, which explains the poor results in terms of edge detectability. Moreover, GraphEM provides a better RMSE score on all examples. Regarding the graph structure, we can observe that GraphEM has also better detection scores when compared with both PGC and CGC. We also display an example of graph reconstruction for dataset C in Fig.~\ref{fig:datasetC}, illustrating the ability of GraphEM to  recover the graph shape and weights. 

\begin{table}[t]
\centering
{\small
\begin{tabular}{|c|c||c|c|c|c|c|c|}
\cline{2-8}
\multicolumn{1}{c|}{ } & method & RMSE & accur. & prec. & recall & spec. & F1\\
\hline
\multirow{4}{*}{A} & GraphEM & $0.081$ & $0.9104$ & $0.9880$ & $0.7407$ & $0.9952$ & $0.8463$\\
  & MLEM & $0.149$ & $0.3333$ & $0.3333$ & $1$ & $0$ & $0.5$\\
  & PGC& - & $0.8765$ & $0.9474$ & $0.6667$ & $0.9815$& $0.7826$\\ 
  & CGC & - & $0.8765$ & $1$ & $0.6293$ & $1$& $0.7727$\\ 
\hline
\multirow{4}{*}{B} & GraphEM & $0.082$ & $0.9113$ & $0.9914$ & $ 0.7407$ & $0.9967$ & $0.8477$\\
& MLEM & $0.148$ & $0.3333$ & $0.3333$ & $1$ & $0$ & $0.5$\\
  & PGC & - & $0.8889$ & $1$ & $0.6667$ & $1$& $0.8$\\ 
  & CGC & - &$0.8889$ & $1$ & $0.6667$ & $1$& $0.8$\\ 
\hline
\multirow{4}{*}{C} & GraphEM & $0.120$ & $0.9231$& $0.9401$ & $0.77$ & $0.9785$ & $0.8427$\\
& MLEM & $0.238$ & $0.2656$ & $0.2656$ & $1$ & $0$ & $0.4198$\\
  & PGC& - & $0.9023$ & $0.9778$ & $0.6471$ & $0.9949$& $0.7788$\\ 
  & CGC& - & $0.8555$ & $0.9697$ & $0.4706$ & $0.9949$& $0.6337$\\ 
\hline
\multirow{4}{*}{D} & GraphEM& $0.121$ & $0.9247$ & $0.9601$ & $0.7547$ & $0.9862$ & $0.8421$\\
  & MLEM& $0.239$ & $0.2656$ & $0.2656$ & $1$ & $0$ & $0.4198$\\
  & PGC & - & $0.8906$ & $0.9$ & $0.6618$ & $0.9734$& $0.7627$\\ 
    & CGC& - & $0.8477$ & $0.9394$ & $0.4559$ & $0.9894$& $0.6139$\\ 
\hline
\end{tabular}
}

\caption{Results for GraphEM, MLEM, PGC and CGC.}
\label{tab:results}
\end{table}

\subsection{Climate modelling and understanding}

Studying the complex inter-dependencies in climate processes is a critical challenge. We use climate models and analytical data to gain insight through observational causal discovery \citep{Runge19natcom}. We evaluate graphEM and compare it to other methods on climate-related problems, such as linking ENSO and the North Atlantic Oscillation.

We used pre-industrial control runs from \cite{Eyring2016}, followed the methodology in \cite{pmlr-v123-runge20a}, and selected $15$ climate variables (hfls, hfss, huss, rlds, rlus, rlut, ta, tas, tasmax, tasmin, uas, va, vas, wap, zg). We de-seasonalized the time series and applied Varimax-rotated PCA to monthly averages. Then used the obtained weights to generate daily component time series, averaging to a $5$-day resolution. 
We randomly picked $N$ component time series, standardised them, and constructed the ground truth by fitting a VAR model as in \cite{pmlr-v123-runge20a}. 
Ground truth was obtained by thresholding coefficients $<0.22$ and keeping random draws of $N$ components with $L=N$ links. We then constructed experiments scenarios modelling climate \cite{pmlr-v123-runge20a}. We use linear models and two sample sizes ($T\approx 100-250$ and $N=5,40$) and accounted for non-stationarity and observational noise, resulting in 12 experiments with $200$ realisations each. We implemented GraphEM in our \href{http://causeme.net}{CAUSEME} web platform and compared the performance to standard algorithms for time series data: VAR \cite{toda1991vector}, GC \cite{granger_investigating_1969}, and PCMCI \cite{Runge19natcom}. Results in Table \ref{tab:results} show that the GraphEM methods outperform VAR and GC in all performance metrics (accuracy, precision, recall, specificity and F1) and PCMCI in recall and overall F1 score. Figure \ref{fig:climateresults} shows the obtained graphs for a particular example and evidences the good detection of links $2, 4\to 5$ unlike with PCMCI, as well as the much sparser (and less convoluted) solution attained compared to VAR and GC. 

\begin{table}[t]
\centering
\setlength{\tabcolsep}{1pt}
{\small
\begin{tabular}{|l|c||c|c|c|c|c|}
\hline
method & best hyperparameters & accur. & prec. & recall & spec. & F1\\
\hline
GraphEM \cite{elvira2022graphem} & 
$\sigma_{\bf R}=0.1, \sigma_{\bf P}=10^{-4}$, $\gamma_{1} = 50$  & ${\bf 0.72}$ & $0.75$ & ${\bf 0.55}$ & $0.86$ & ${\bf 0.63}$\\
VAR \cite{toda1991vector} & $\ell=8$ & $0.56$ & $0.50$ & $0.46$ & $0.64$ & $0.48$\\
Granger \cite{granger_investigating_1969} & $\ell=8$ & $0.6$ & $0.57$ & $0.36$ & $0.79$& $0.44$\\ 
PCMCI \cite{Runge19natcom} & $\tau_{\text{max}}=8$, $\alpha_{\text{PC}}=0.05$, ParCorr & {\bf $0.72$} & ${\bf 0.83}$ & $0.45$ & ${\bf 0.93}$ & $0.59$\\ 

\hline

\end{tabular}
}
\caption{Results for GraphEM \cite{elvira2022graphem} (Lasso), VAR \cite{toda1991vector}, GC \cite{granger_investigating_1969} and PCMCI \cite{Runge19natcom} for a set of climate problems. Best results are highlighted in bold.}
\label{tab:results}
\end{table}

\begin{figure*}[t!]
\centering
\setlength{\tabcolsep}{-3pt}
\begin{tabular}{cccccc}
   & Truth & VAR & Granger & PCMCI & GraphEM  \\
    \includegraphics[width=3cm]{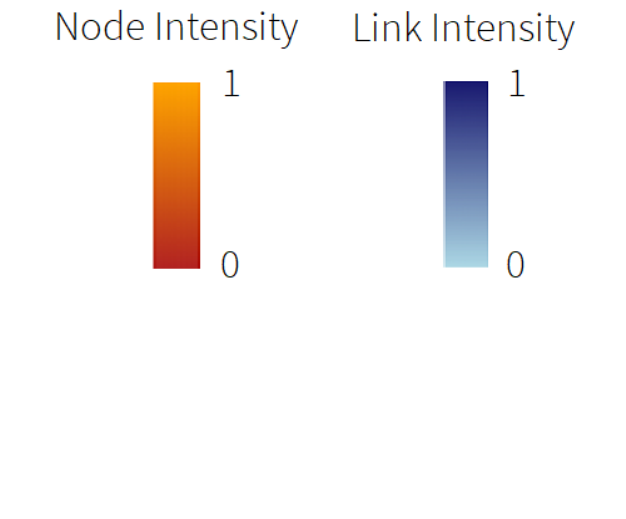} &
    \includegraphics[width=3.5cm]{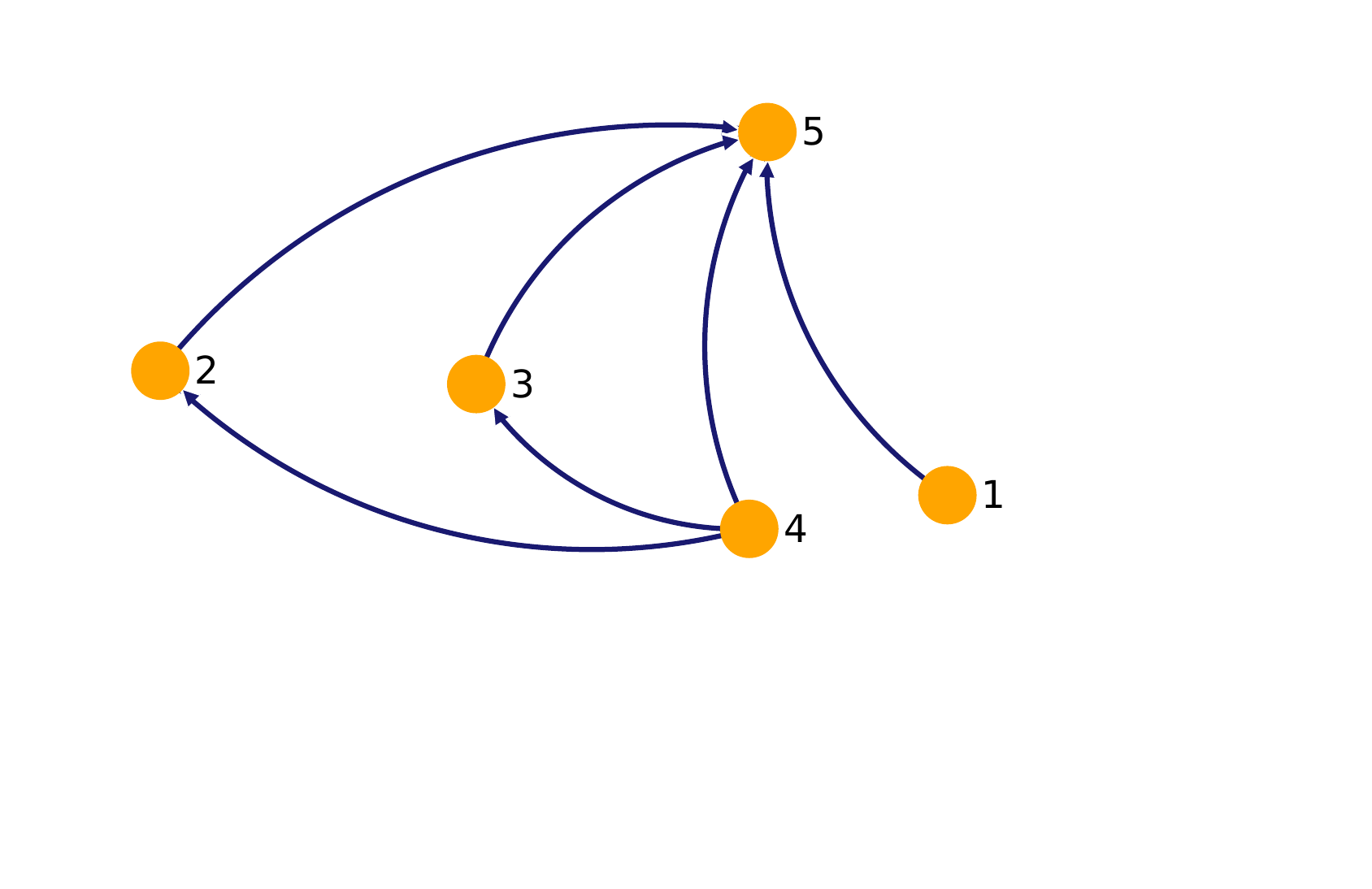} & 
\hspace{-1cm}\includegraphics[width=3.5cm]{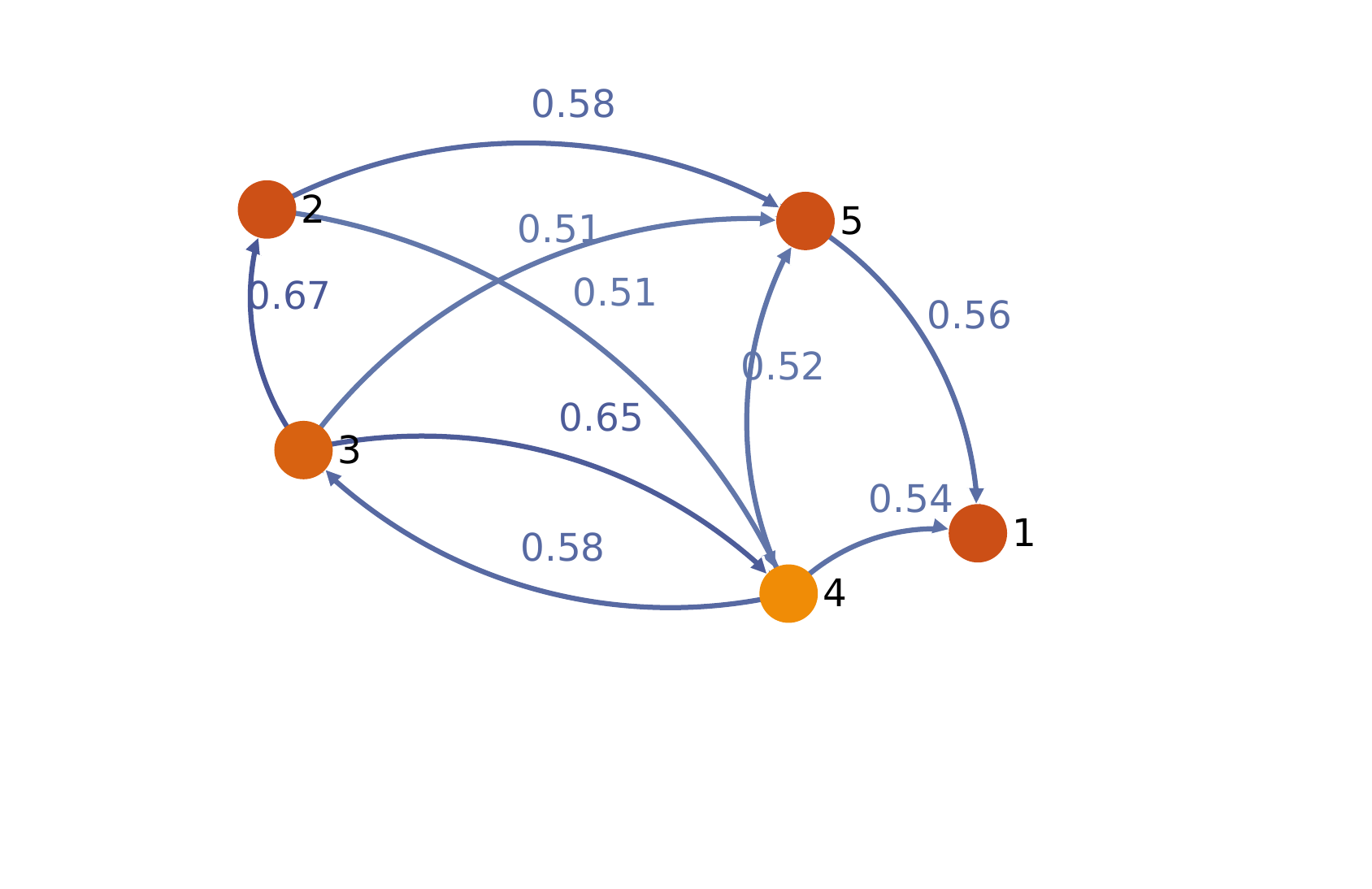} & 
    \includegraphics[width=3.5cm]{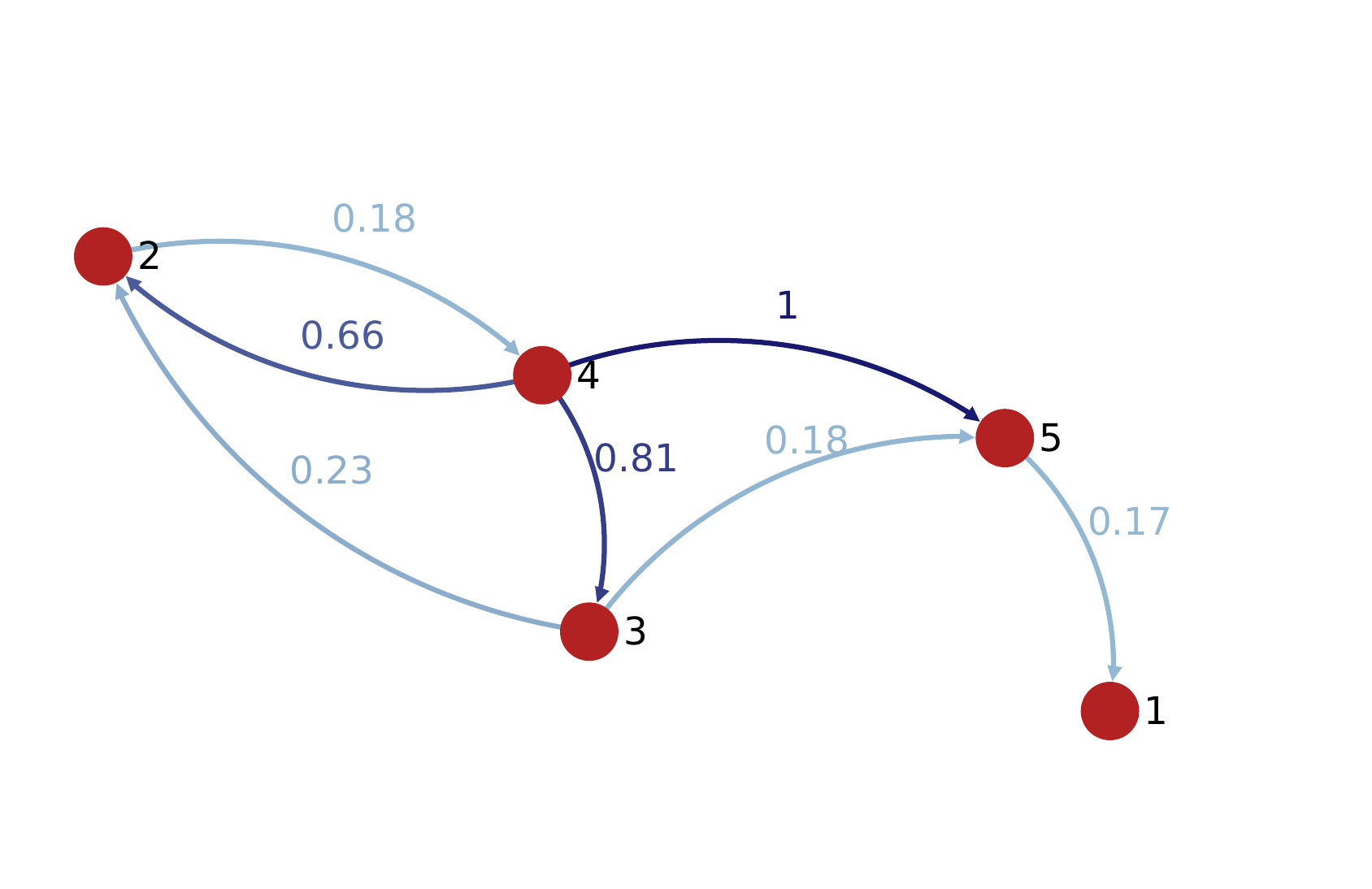} &
    \includegraphics[width=3.5cm]{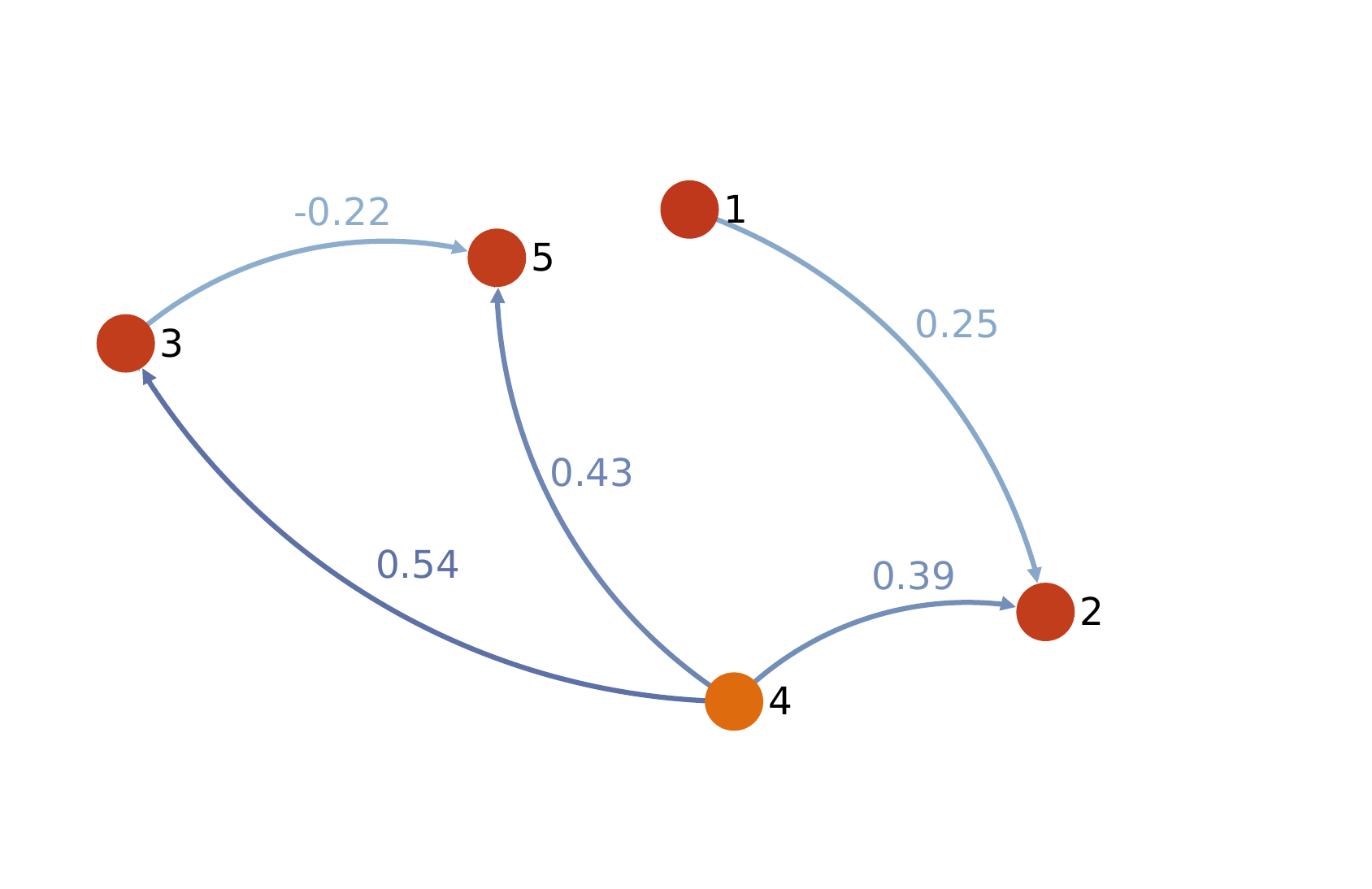} & 
    \includegraphics[width=3.5cm]{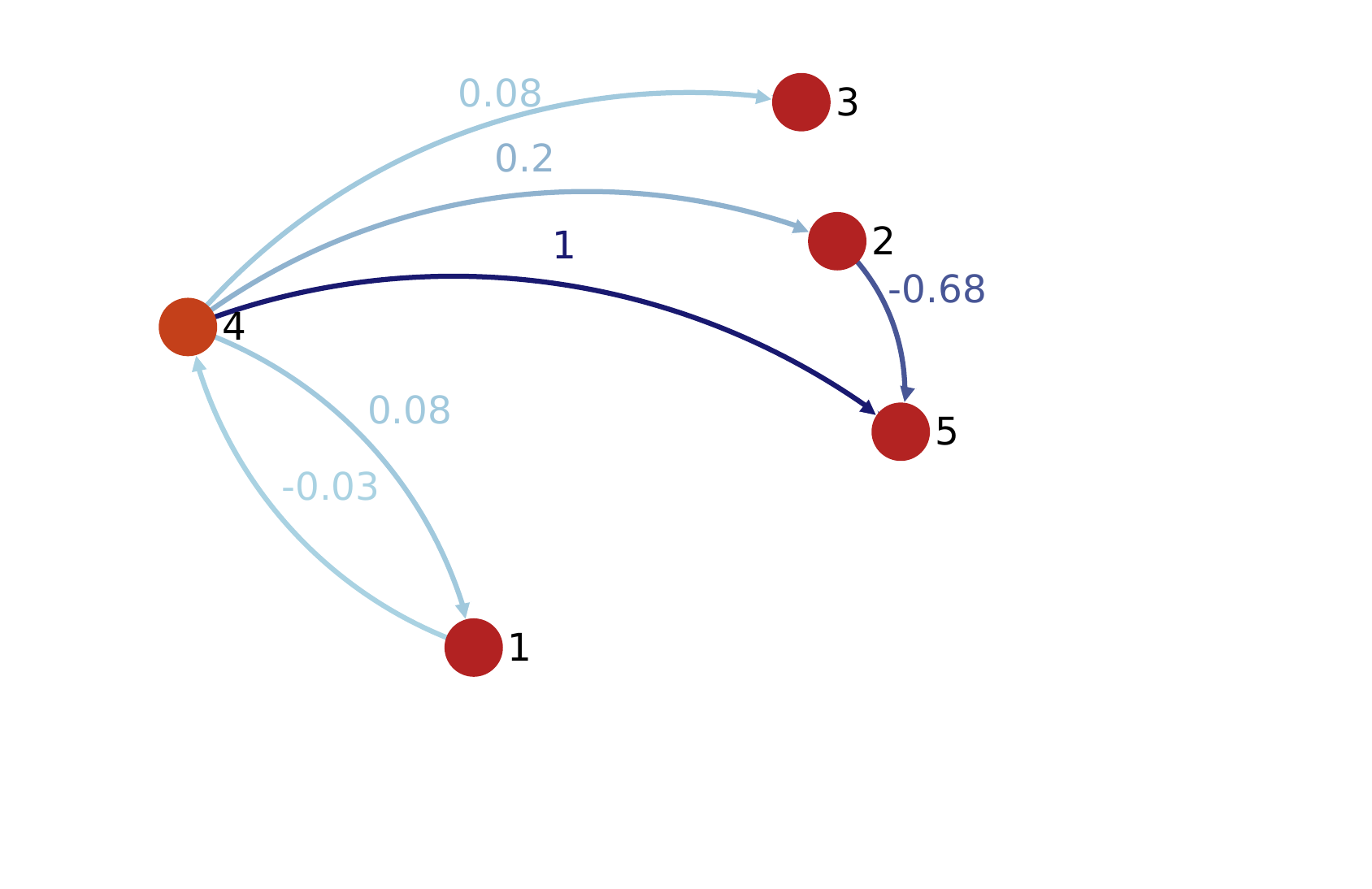}
\end{tabular}
\vspace{-0.7cm}
\caption{Causal graphs extracted by different methods in an illustrative climate problem.}    \label{fig:climateresults}
\end{figure*}

\section{CONCLUSION}
\label{sec:conclusion}

GraphEM is an EM method for estimating the linear operator in linear-Gaussian state-space models. GraphEM incorporates a sparsity constraint using a Lasso penalty term, which makes it well suited for modelling the state entries interactions as a compact and interpretable graph. The inner minimisation problem is solved using a proximal splitting algorithm. Numerical results show that GraphEM outperforms several other techniques for graphical modelling. Results in more challenging real problems in the Earth, Social and Climate sciences will be illustrated at the time of the conference.

\section{Acknowledgements} The work of V. E. is supported by the \emph{Agence Nationale de la Recherche} of France under PISCES (ANR-17-CE40-0031-01), the Leverhulme Research Fellowship (RF-2021-593), and by ARL/ARO under grants W911NF-20-1-0126 and W911NF-22-1-0235. \'E.C. acknowledges support from the European
Research Council (ERC) Starting Grant MAJORIS ERC-2019-STG-850925, and G.C-V. acknowledges support from the ERC Synergy Grant USMILE (grant agreement 855187), the Fundaci\'on BBVA with the project `Causal inference in the human-biosphere coupled system (SCALE)', and the Microsoft Climate Research Initiative through the Causal4Africa project.

\bibliographystyle{ACM-Reference-Format}
\bibliography{references}


\begin{thebibliography}{33}


\ifx \showCODEN    \undefined \def \showCODEN     #1{\unskip}     \fi
\ifx \showDOI      \undefined \def \showDOI       #1{#1}\fi
\ifx \showISBNx    \undefined \def \showISBNx     #1{\unskip}     \fi
\ifx \showISBNxiii \undefined \def \showISBNxiii  #1{\unskip}     \fi
\ifx \showISSN     \undefined \def \showISSN      #1{\unskip}     \fi
\ifx \showLCCN     \undefined \def \showLCCN      #1{\unskip}     \fi
\ifx \shownote     \undefined \def \shownote      #1{#1}          \fi
\ifx \showarticletitle \undefined \def \showarticletitle #1{#1}   \fi
\ifx \showURL      \undefined \def \showURL       {\relax}        \fi
\providecommand\bibfield[2]{#2}
\providecommand\bibinfo[2]{#2}
\providecommand\natexlab[1]{#1}
\providecommand\showeprint[2][]{arXiv:#2}

\bibitem[Bach et~al\mbox{.}(2012)]%
        {Bach}
\bibfield{author}{\bibinfo{person}{F. Bach}, \bibinfo{person}{R. Jenatton},
  \bibinfo{person}{J. Mairal}, {and} \bibinfo{person}{G. Obozinski}.}
  \bibinfo{year}{2012}\natexlab{}.
\newblock \showarticletitle{Optimization with Sparsity-Inducing Penalties}.
\newblock \bibinfo{journal}{\emph{Foundations Trends Machine Learning}}
  \bibinfo{volume}{4}, \bibinfo{number}{1} (\bibinfo{date}{Jan.}
  \bibinfo{year}{2012}), \bibinfo{pages}{1--106}.
\newblock
\showISSN{1935-8237}


\bibitem[{Bach} and {Jordan}(2004)]%
        {Bach04}
\bibfield{author}{\bibinfo{person}{F.~R. {Bach}} {and} \bibinfo{person}{M.~I.
  {Jordan}}.} \bibinfo{year}{2004}\natexlab{}.
\newblock \showarticletitle{Learning graphical models for stationary time
  series}.
\newblock \bibinfo{journal}{\emph{IEEE Transactions on Signal Processing}}
  \bibinfo{volume}{52}, \bibinfo{number}{8} (\bibinfo{date}{Aug.}
  \bibinfo{year}{2004}), \bibinfo{pages}{2189--2199}.
\newblock


\bibitem[{Barber} and {Cemgil}(2010)]%
        {Barber10}
\bibfield{author}{\bibinfo{person}{D. {Barber}} {and} \bibinfo{person}{A.~T.
  {Cemgil}}.} \bibinfo{year}{2010}\natexlab{}.
\newblock \showarticletitle{Graphical Models for Time-Series}.
\newblock \bibinfo{journal}{\emph{IEEE Signal Processing Magazine}}
  \bibinfo{volume}{27}, \bibinfo{number}{6} (\bibinfo{date}{Nov}
  \bibinfo{year}{2010}), \bibinfo{pages}{18--28}.
\newblock


\bibitem[Barnett and Seth(2015)]%
        {barnett2015granger}
\bibfield{author}{\bibinfo{person}{L Barnett} {and} \bibinfo{person}{A.K.
  Seth}.} \bibinfo{year}{2015}\natexlab{}.
\newblock \showarticletitle{Granger causality for state-space models}.
\newblock \bibinfo{journal}{\emph{Physical Review E}} \bibinfo{volume}{91},
  \bibinfo{number}{4} (\bibinfo{year}{2015}), \bibinfo{pages}{040101}.
\newblock


\bibitem[Bauschke and Combettes(2017)]%
        {Bauschke}
\bibfield{author}{\bibinfo{person}{H.~H. Bauschke} {and} \bibinfo{person}{P.~L.
  Combettes}.} \bibinfo{year}{2017}\natexlab{}.
\newblock \bibinfo{booktitle}{\emph{Convex Analysis and Monotone Operator
  Theory in Hilbert Spaces}}.
\newblock \bibinfo{publisher}{Springer, New York}.
\newblock
\newblock
\shownote{2nd edition}.


\bibitem[Benfenati et~al\mbox{.}(2018)]%
        {Benfenati18}
\bibfield{author}{\bibinfo{person}{A. Benfenati}, \bibinfo{person}{E.
  Chouzenoux}, {and} \bibinfo{person}{J.-C. Pesquet}.}
  \bibinfo{year}{2018}\natexlab{}.
\newblock \bibinfo{booktitle}{\emph{A Proximal Approach for a Class of Matrix
  Optimization Problems}}.
\newblock \bibinfo{type}{{T}echnical {R}eport}.
\newblock


\bibitem[Bressler and Seth(2011)]%
        {Bressler2011}
\bibfield{author}{\bibinfo{person}{S.~L. Bressler} {and} \bibinfo{person}{A.~K.
  Seth}.} \bibinfo{year}{2011}\natexlab{}.
\newblock \showarticletitle{{W}iener--{G}ranger causality: a well established
  methodology}.
\newblock \bibinfo{journal}{\emph{NeuroImage}} \bibinfo{volume}{58},
  \bibinfo{number}{2} (\bibinfo{year}{2011}), \bibinfo{pages}{323--329}.
\newblock


\bibitem[Chouzenoux and Elvira(2020)]%
        {chouzenoux2020}
\bibfield{author}{\bibinfo{person}{E. Chouzenoux} {and} \bibinfo{person}{V.
  Elvira}.} \bibinfo{year}{2020}\natexlab{}.
\newblock \showarticletitle{{GraphEM: EM} Algorithm for Blind {K}alman
  Filtering under Graphical Sparsity Constraints}. In
  \bibinfo{booktitle}{\emph{ICASSP}}. \bibinfo{pages}{5840--5844}.
\newblock


\bibitem[Combettes and Pesquet(2007)]%
        {CombettesDR}
\bibfield{author}{\bibinfo{person}{P.L. Combettes} {and} \bibinfo{person}{J.C.
  Pesquet}.} \bibinfo{year}{2007}\natexlab{}.
\newblock \showarticletitle{A {Douglas-Rachford} splitting approach to
  nonsmooth convex variational signal recovery}.
\newblock \bibinfo{journal}{\emph{IEEE Journal of Selected Topics in Signal
  Processing}} \bibinfo{volume}{1}, \bibinfo{number}{4} (\bibinfo{date}{Dec.}
  \bibinfo{year}{2007}), \bibinfo{pages}{564--574}.
\newblock


\bibitem[Combettes and Pesquet(2010)]%
        {Combettes2010}
\bibfield{author}{\bibinfo{person}{P.L. Combettes} {and} \bibinfo{person}{J.C.
  Pesquet}.} \bibinfo{year}{2010}\natexlab{}.
\newblock \bibinfo{booktitle}{\emph{Proximal Splitting Methods in Signal
  Processing}}.
\newblock \bibinfo{publisher}{Springer-Verlag, New York}.
\newblock


\bibitem[Eichler(2012)]%
        {Eichler2012}
\bibfield{author}{\bibinfo{person}{M. Eichler}.}
  \bibinfo{year}{2012}\natexlab{}.
\newblock \showarticletitle{Graphical modelling of multivariate time series}.
\newblock \bibinfo{journal}{\emph{Probability Theory and Related Fields}}
  \bibinfo{volume}{153}, \bibinfo{number}{1} (\bibinfo{date}{01 Jun.}
  \bibinfo{year}{2012}), \bibinfo{pages}{233--268}.
\newblock
\showISSN{1432-2064}


\bibitem[Elvira and Chouzenoux(2022)]%
        {elvira2022graphem}
\bibfield{author}{\bibinfo{person}{V. Elvira} {and} \bibinfo{person}{E.
  Chouzenoux}.} \bibinfo{year}{2022}\natexlab{}.
\newblock \showarticletitle{Graphical Inference in Linear-{G}aussian
  State-Space Models}.
\newblock \bibinfo{journal}{\emph{IEEE Transactions on Signal Processing}}
  \bibinfo{volume}{70} (\bibinfo{year}{2022}), \bibinfo{pages}{4757--4771}.
\newblock


\bibitem[Eyring et~al\mbox{.}(2016)]%
        {Eyring2016}
\bibfield{author}{\bibinfo{person}{V. Eyring}, \bibinfo{person}{S. Bony},
  \bibinfo{person}{G.A. Meehl}, \bibinfo{person}{C.A. Senior},
  \bibinfo{person}{Bjorn Stevens}, \bibinfo{person}{R.J. Stouffer}, {and}
  \bibinfo{person}{K.E. Taylor}.} \bibinfo{year}{2016}\natexlab{}.
\newblock \showarticletitle{{Overview of the Coupled Model Intercomparison
  Project Phase 6 (CMIP6) experimental design and organization}}.
\newblock \bibinfo{journal}{\emph{Geoscientific Model Development}}
  \bibinfo{volume}{9}, \bibinfo{number}{5} (\bibinfo{year}{2016}),
  \bibinfo{pages}{1937--1958}.
\newblock


\bibitem[Granger(1969)]%
        {granger_investigating_1969}
\bibfield{author}{\bibinfo{person}{C.~W.~J. Granger}.}
  \bibinfo{year}{1969}\natexlab{}.
\newblock \showarticletitle{Investigating Causal Relations by Econometric
  Models and Cross-spectral Methods}.
\newblock \bibinfo{journal}{\emph{Econometrica}} \bibinfo{volume}{37},
  \bibinfo{number}{3} (\bibinfo{year}{1969}), \bibinfo{pages}{424--438}.
\newblock
\showISSN{00129682, 14680262}


\bibitem[Hicks et~al\mbox{.}(1980)]%
        {hicks1980causality}
\bibfield{author}{\bibinfo{person}{J. Hicks} {et~al\mbox{.}}}
  \bibinfo{year}{1980}\natexlab{}.
\newblock \bibinfo{booktitle}{\emph{Causality in economics}}.
\newblock \bibinfo{publisher}{Australian National University Press}.
\newblock


\bibitem[Hyv\"{a}rinen et~al\mbox{.}(2010)]%
        {hyvarinen_estimation_2010}
\bibfield{author}{\bibinfo{person}{A. Hyv\"{a}rinen}, \bibinfo{person}{K.
  Zhang}, \bibinfo{person}{S. Shimizu}, {and} \bibinfo{person}{P.O. Hoyer}.}
  \bibinfo{year}{2010}\natexlab{}.
\newblock \showarticletitle{Estimation of a structural vector autoregression
  model using non-gaussianity}.
\newblock \bibinfo{journal}{\emph{Journal of Machine Learning Research}}
  \bibinfo{volume}{11}, \bibinfo{number}{May} (\bibinfo{year}{2010}),
  \bibinfo{pages}{1709--1731}.
\newblock


\bibitem[Jacobson and Fessler(2007)]%
        {jacobson}
\bibfield{author}{\bibinfo{person}{M.W. Jacobson} {and} \bibinfo{person}{J.A.
  Fessler}.} \bibinfo{year}{2007}\natexlab{}.
\newblock \showarticletitle{An expanded theoretical treatment of
  iteration-dependent majorize-minimize algorithms}.
\newblock \bibinfo{journal}{\emph{IEEE Transactions on Image Processing}}
  \bibinfo{volume}{16}, \bibinfo{number}{10} (\bibinfo{year}{2007}),
  \bibinfo{pages}{2411–2422}.
\newblock


\bibitem[Kalman(1960)]%
        {Kalman60}
\bibfield{author}{\bibinfo{person}{R.~E. Kalman}.}
  \bibinfo{year}{1960}\natexlab{}.
\newblock \showarticletitle{A New Approach to Linear Filtering and Prediction
  Problems}.
\newblock \bibinfo{journal}{\emph{Journal of Basic Engineering}}
  \bibinfo{volume}{82} (\bibinfo{year}{1960}), \bibinfo{pages}{35--45}.
\newblock


\bibitem[Liang(2016)]%
        {liang_information_2015}
\bibfield{author}{\bibinfo{person}{X.~San Liang}.}
  \bibinfo{year}{2016}\natexlab{}.
\newblock \showarticletitle{Information flow and causality as rigorous notions
  ab initio}.
\newblock \bibinfo{journal}{\emph{Phys. Rev. E}}  \bibinfo{volume}{94}
  (\bibinfo{date}{Nov} \bibinfo{year}{2016}), \bibinfo{pages}{052201}.
\newblock
Issue 5.


\bibitem[Luengo et~al\mbox{.}(2019)]%
        {Luengo19}
\bibfield{author}{\bibinfo{person}{D. Luengo}, \bibinfo{person}{G.~Rios-Mu\
  noz}, \bibinfo{person}{V. Elvira}, \bibinfo{person}{C. S\'anchez}, {and}
  \bibinfo{person}{A. Art\'es-Rodr\'iguez}.} \bibinfo{year}{2019}\natexlab{}.
\newblock \showarticletitle{Hierarchical Algorithms for Causality Retrieval in
  Atrial Fibrillation Intracavitary Electrograms}.
\newblock \bibinfo{journal}{\emph{IEEE Jour. Biomed. Hea. Inf.}}
  \bibinfo{volume}{12}, \bibinfo{number}{1} (\bibinfo{date}{Jan}
  \bibinfo{year}{2019}), \bibinfo{pages}{143--155}.
\newblock


\bibitem[Marini and Singer(1988)]%
        {marini1988causality}
\bibfield{author}{\bibinfo{person}{M.M. Marini} {and} \bibinfo{person}{B.
  Singer}.} \bibinfo{year}{1988}\natexlab{}.
\newblock \showarticletitle{Causality in the social sciences}.
\newblock \bibinfo{journal}{\emph{Sociological methodology}}
  \bibinfo{volume}{18} (\bibinfo{year}{1988}), \bibinfo{pages}{347--409}.
\newblock


\bibitem[Massey(1990)]%
        {massey1990causality}
\bibfield{author}{\bibinfo{person}{J. Massey}.}
  \bibinfo{year}{1990}\natexlab{}.
\newblock \showarticletitle{Causality, feedback and directed information}. In
  \bibinfo{booktitle}{\emph{Proc. Int. Symp. Inf. Theory Applic.(ISITA-90)}}.
  Citeseer, \bibinfo{pages}{303--305}.
\newblock


\bibitem[Peters et~al\mbox{.}(2013)]%
        {peters_causal_2013}
\bibfield{author}{\bibinfo{person}{J. Peters}, \bibinfo{person}{D. Janzing},
  {and} \bibinfo{person}{B. Sch{\"o}lkopf}.} \bibinfo{year}{2013}\natexlab{}.
\newblock \showarticletitle{Causal inference on time series using restricted
  structural equation models}. In \bibinfo{booktitle}{\emph{Advances in
  {Neural} {Information} {Processing} {Systems}}}. \bibinfo{pages}{154--162}.
\newblock


\bibitem[Reid et~al\mbox{.}(2019)]%
        {reid2019advancing}
\bibfield{author}{\bibinfo{person}{A.T. Reid}, \bibinfo{person}{D.B. Headley},
  \bibinfo{person}{R.D. Mill}, \bibinfo{person}{R. Sanchez-Romero},
  \bibinfo{person}{L.Q. Uddin}, \bibinfo{person}{D. Marinazzo},
  \bibinfo{person}{D.~J Lurie}, \bibinfo{person}{P.A. Vald{\'e}s-Sosa},
  \bibinfo{person}{S.J. Hanson}, {and} \bibinfo{person}{B.B. Biswal}.}
  \bibinfo{year}{2019}\natexlab{}.
\newblock \showarticletitle{Advancing functional connectivity research from
  association to causation}.
\newblock \bibinfo{journal}{\emph{Nature Neuroscience}} \bibinfo{volume}{1},
  \bibinfo{number}{10} (\bibinfo{year}{2019}).
\newblock


\bibitem[Runge et~al\mbox{.}(2019)]%
        {Runge19natcom}
\bibfield{author}{\bibinfo{person}{J. Runge}, \bibinfo{person}{S. Bathiany},
  \bibinfo{person}{E. Bollt}, \bibinfo{person}{G. Camps-Valls},
  \bibinfo{person}{D. Coumou}, \bibinfo{person}{E. Deyle}, \bibinfo{person}{C.
  Clymour}, \bibinfo{person}{M. Kretschmer}, \bibinfo{person}{M. Mahecha},
  \bibinfo{person}{J. Mu{\~n}oz-Mar\'i}, \bibinfo{person}{E. van Nes},
  \bibinfo{person}{J. Peters}, \bibinfo{person}{R. Quax}, \bibinfo{person}{M.
  Reichstein}, \bibinfo{person}{M. Scheffer}, \bibinfo{person}{B. Sch\"olkopf},
  \bibinfo{person}{P. Spirtes}, \bibinfo{person}{G. Sugihara},
  \bibinfo{person}{J. Sun}, \bibinfo{person}{K. Zhang}, {and}
  \bibinfo{person}{J. Zscheischler}.} \bibinfo{year}{2019}\natexlab{}.
\newblock \showarticletitle{Inferring causation from time series with
  perspectives in {E}arth system sciences}.
\newblock \bibinfo{journal}{\emph{Nature Communications}} \bibinfo{volume}{10},
  \bibinfo{number}{2553} (\bibinfo{year}{2019}).
\newblock


\bibitem[Runge et~al\mbox{.}(2020)]%
        {pmlr-v123-runge20a}
\bibfield{author}{\bibinfo{person}{J. Runge}, \bibinfo{person}{X.-A. Tibau},
  \bibinfo{person}{M. Bruhns}, \bibinfo{person}{J. Mu\~{n}oz Mar\'{i}}, {and}
  \bibinfo{person}{G. Camps-Valls}.} \bibinfo{year}{2020}\natexlab{}.
\newblock \showarticletitle{The Causality for Climate Competition}. In
  \bibinfo{booktitle}{\emph{NeurIPS 2019 Competition and Demonstration Track}},
  \bibfield{editor}{\bibinfo{person}{Hugo~Jair Escalante} {and}
  \bibinfo{person}{Raia Hadsell}} (Eds.), Vol.~\bibinfo{volume}{123}.
  \bibinfo{publisher}{PMLR}, \bibinfo{pages}{110--120}.
\newblock


\bibitem[Sarkka(2013)]%
        {Sarkka}
\bibfield{author}{\bibinfo{person}{S. Sarkka}.}
  \bibinfo{year}{2013}\natexlab{}.
\newblock \bibinfo{booktitle}{\emph{Bayesian Filtering and Smoothing}
  (\bibinfo{edition}{3} ed.)}.
\newblock


\bibitem[Schmidt et~al\mbox{.}(2007)]%
        {Schmidt}
\bibfield{author}{\bibinfo{person}{M. Schmidt}, \bibinfo{person}{G. Fung},
  {and} \bibinfo{person}{R. Rosales}.} \bibinfo{year}{2007}\natexlab{}.
\newblock \showarticletitle{Fast Optimization Methods for L1 Regularization: A
  Comparative Study and Two New Approaches}. In
  \bibinfo{booktitle}{\emph{Machine Learning: ECML 2007}},
  \bibfield{editor}{\bibinfo{person}{J.~N. Kok},
  \bibinfo{person}{J.~Koronacki}, \bibinfo{person}{R.L. Mantaras},
  \bibinfo{person}{S.~Matwin}, \bibinfo{person}{D.~Mladeni{\v{c}}}, {and}
  \bibinfo{person}{A.~Skowron}} (Eds.). \bibinfo{publisher}{Springer Berlin
  Heidelberg}, \bibinfo{address}{Berlin, Heidelberg},
  \bibinfo{pages}{286--297}.
\newblock
\showISBNx{978-3-540-74958-5}


\bibitem[Schreiber(2000)]%
        {schreiber_measuring_2000}
\bibfield{author}{\bibinfo{person}{Thomas Schreiber}.}
  \bibinfo{year}{2000}\natexlab{}.
\newblock \showarticletitle{Measuring Information Transfer}.
\newblock \bibinfo{journal}{\emph{Phys. Rev. Lett.}}  \bibinfo{volume}{85}
  (\bibinfo{date}{Jul} \bibinfo{year}{2000}), \bibinfo{pages}{461--464}.
\newblock
Issue 2.


\bibitem[Sugihara et~al\mbox{.}(2012)]%
        {sugihara_detecting_2012}
\bibfield{author}{\bibinfo{person}{G. Sugihara}, \bibinfo{person}{R. May},
  \bibinfo{person}{H. Ye}, \bibinfo{person}{C.-h. Hsieh}, \bibinfo{person}{E.
  Deyle}, \bibinfo{person}{M. Fogarty}, {and} \bibinfo{person}{S. Munch}.}
  \bibinfo{year}{2012}\natexlab{}.
\newblock \showarticletitle{Detecting Causality in Complex Ecosystems}.
\newblock \bibinfo{journal}{\emph{Science}} \bibinfo{volume}{338},
  \bibinfo{number}{6106} (\bibinfo{year}{2012}), \bibinfo{pages}{496--500}.
\newblock


\bibitem[Tibshirani(1996)]%
        {Tibshirani}
\bibfield{author}{\bibinfo{person}{R. Tibshirani}.}
  \bibinfo{year}{1996}\natexlab{}.
\newblock \showarticletitle{Regression Shrinkage and Selection via the Lasso}.
\newblock \bibinfo{journal}{\emph{Journal of the Royal Statistical Society.
  Series B (Methodological)}} \bibinfo{volume}{58}, \bibinfo{number}{1}
  (\bibinfo{year}{1996}), \bibinfo{pages}{267--288}.
\newblock
\showISSN{00359246}


\bibitem[Toda(1991)]%
        {toda1991vector}
\bibfield{author}{\bibinfo{person}{H. Toda}.} \bibinfo{year}{1991}\natexlab{}.
\newblock \bibinfo{booktitle}{\emph{Vector autoregression and causality}}.
\newblock \bibinfo{publisher}{Yale University}.
\newblock


\bibitem[White et~al\mbox{.}(2011)]%
        {white_linking_2011}
\bibfield{author}{\bibinfo{person}{H. White}, \bibinfo{person}{K. Chalak},
  \bibinfo{person}{X. Lu}, {and} \bibinfo{person}{{others}}.}
  \bibinfo{year}{2011}\natexlab{}.
\newblock \showarticletitle{Linking {Granger} {Causality} and the {Pearl}
  {Causal} {Model} with {Settable} {Systems}.}. In
  \bibinfo{booktitle}{\emph{{NIPS} {Mini}-{Symposium} on {Causality} in {Time}
  {Series}}}. \bibinfo{pages}{1--29}.
\newblock


\end{thebibliography}

\end{document}